\documentclass{article}

\usepackage{arxiv}

\usepackage[utf8]{inputenc} 
\usepackage[T1]{fontenc}    
\usepackage{hyperref}       
\usepackage{url}            
\usepackage{booktabs}       
\usepackage{amsfonts}       
\usepackage{nicefrac}       
\usepackage{microtype}      
\usepackage{lipsum}
\usepackage{graphicx}
\graphicspath{ {./images/} }

\usepackage[utf8]{inputenc} 
\usepackage[T1]{fontenc}    
\usepackage{hyperref}       
\usepackage{url}            
\usepackage{booktabs}       
\usepackage{amsfonts}       
\usepackage{nicefrac}       
\usepackage{microtype}      
\usepackage{xcolor}         

\usepackage[utf8]{inputenc} 
\usepackage[T1]{fontenc}    
\usepackage{hyperref}       
\usepackage{url}            
\usepackage{booktabs}       
\usepackage{amsfonts}       
\usepackage{nicefrac}       
\usepackage{microtype}      
\usepackage{xcolor}         
\usepackage{bm}
\usepackage{amsfonts,amssymb}
\usepackage{amsmath}
\usepackage{enumerate}
\usepackage{multirow}
\usepackage{subfigure}
\usepackage{epstopdf}
\usepackage{graphics}
\usepackage{epsfig}
\usepackage{enumitem}
\usepackage{url}
\usepackage{bbm}
\usepackage{times}
\usepackage{soul}
\usepackage{url}
\usepackage[small]{caption}
\usepackage{graphicx}
\usepackage{amsmath}
\usepackage{booktabs}
\usepackage{algorithm}
\usepackage{algorithmic}
\usepackage{enumitem}
\usepackage{amsmath}
\usepackage{amssymb}
\usepackage{mathtools}
\usepackage{amsthm}
\usepackage{times}
\usepackage{soul}
\usepackage{url}
\usepackage{times}
\usepackage{soul}
\usepackage{url}
\usepackage{bm}
\usepackage{subfigure}

\newcommand{\plcr}[0]{\textsc{Plcr}}
\newcommand{\pico}[0]{\textsc{Pico}}
\newcommand{\cavl}[0]{\textsc{Cavl}}
\newcommand{\lws}[0]{\textsc{Lws}}
\newcommand{\vpll}[0]{\textsc{Valen}}
\newcommand{\rcpll}[0]{\textsc{RC}}
\newcommand{\ccpll}[0]{\textsc{CC}}
\newcommand{\proden}[0]{\textsc{Proden}}

\newcommand{\plable}[0]{\textsc{Able}}
\newcommand{\plidgp}[0]{\textsc{Idgp}}
\newcommand{\plpop}[0]{\textsc{Pop}}
\newcommand{\pldirk}[0]{\textsc{Dirk}}
\newcommand{\plsdct}[0]{\textsc{Sdct}}

\newcommand{\ours}[0]{\textsc{Rplg}}
\newcommand{\oursnm}[0]{\textsc{Rplg}-\textsc{Nm}}
\newtheorem{theorem}{Theorem}

\newtheorem{assumption}{Assumption}

\newtheorem{definition}{Definition}

\title{Reduction-based Pseudo-label Generation for Instance-dependent Partial Label Learning}

\author{
 Congyu Qiao \\
  School of Computer Science and Engineering\\
  Southeast University\\
  Nanjing, China 211189 \\
  \texttt{qiaocy@seu.edu.cn} \\
   \And
 Ning Xu \\
  School of Computer Science and Engineering\\
  Southeast University\\
  Nanjing, China 211189 \\
  \texttt{xning@seu.edu.cn} \\
  \And
 Yihao Hu \\
  School of Computer Science and Engineering\\
  Southeast University\\
  Nanjing, China 211189 \\
  \texttt{yhhu@seu.edu.cn} \\
  \And
  Xin Geng \\
  School of Computer Science and Engineering\\
  Southeast University\\
  Nanjing, China 211189 \\
  \texttt{xgeng@seu.edu.cn} \\
}

\begin{document}
\maketitle
\begin{abstract}
\textit{Instance-dependent Partial Label Learning} (ID-PLL) aims to learn a multi-class predictive model given training instances annotated with candidate labels related to features, among which correct labels are hidden fixed but unknown. The previous works involve leveraging the identification capability of the training model itself to iteratively refine supervision information. However, these methods overlook a critical aspect of ID-PLL: the training model is prone to overfitting on incorrect candidate labels, thereby providing poor supervision information and creating a bottleneck in training. In this paper, we propose to leverage reduction-based pseudo-labels to alleviate the influence of incorrect candidate labels and train our predictive model to overcome this bottleneck. Specifically, reduction-based pseudo-labels are generated by performing weighted aggregation on the outputs of a multi-branch auxiliary model, with each branch trained in a label subspace that excludes certain labels. This approach ensures that each branch explicitly avoids the disturbance of the excluded labels, allowing the pseudo-labels provided for instances troubled by these excluded labels to benefit from the unaffected branches. Theoretically, we demonstrate that reduction-based pseudo-labels exhibit greater consistency with the Bayes optimal classifier compared to pseudo-labels directly generated from the predictive model.
\end{abstract}
\section{Introduction}
Partial Label Learning (PLL), a typical weakly supervised learning paradigm, aims to build a predictive model that assigns the correct label to unseen instances by learning from training instances annotated with a candidate label set that obscures the exact correct label \cite{cour2011learning,chen2014ambiguously,yu2016maximum}. The necessity to learn from such weak supervision naturally arises in ecoinformatics \cite{liu2012conditional,tang2017confidence}, web mining \cite{luo2010learning}, multimedia content analysis \cite{zeng2013learning,chen2017learning}, and other domains, owing to the difficulty in collecting large-scale high-quality datasets.

PLL has been studied along two different routes: the identification-based route\cite{jin2002learning,nguyen2008classification,liu2012conditional,chen2014ambiguously,yu2016maximum, lv2020progressive, feng2020provably, zhang2021exploiting, wang2022pico, wu2022revisiting}, which treats correct labels as the latent variable and tries to identifies them, and the average-based route \cite{hullermeier2006learning,cour2011learning,zhang2015solving,lv2021robustness}, which treats all candidate labels equally. To facilitate practical PLL algorithms, some researchers have focused on instance-dependent PLL (ID-PLL), where incorrect labels related to features are likely to be selected as candidate labels, and tackled this challenge by following the identification-based route. \cite{xu2021instance} explicitly estimate the label distribution aligned with the model output on candidate labels through variational inference. \cite{xia2022ambiguity} induce a contrastive learning framework from ambiguity to refine the representation of the model. \cite{qiao2023decompose} model the generation process of instance-dependent candidate labels by leveraging prior knowledge in the model output \cite{xu2023progressive} propose a theoretically-guaranteed method that progressively identifies incorrect candidate labels by leveraging the margin between the model output values on candidate labels.

These previous approaches involve leveraging the identification capability of the training model itself to iteratively refine supervision information. They exercise this capability through the outputs \cite{xu2021instance, qiao2023decompose, he2023candidate,wu2024distilling} or representations \cite{xia2022ambiguity} from the training model. However, these methods overlook a critical issue in ID-PLL: the training model is prone to overfitting on incorrect labels that are strongly associated with features. As a result, the model generates increasing amounts of incorrect identification information, further leading to the training bottleneck. Unlike previous approaches that iteratively train the predictive model under the influence of misleading supervision before performing identification, we prioritize eliminating the influence of misleading candidate labels, and then train an auxiliary model free from this influence to provide accurate identification information.

In this paper, we propose to utilize reduction-based pseudo-labels to mitigate the influence of incorrect labels and train our predictive model. Specifically, inspired by \cite{SheHSWS021}, reduction-based pseudo-labels are generated by aggregating the outputs of a multi-branch auxiliary model, with each branch trained in a label subspace that excludes certain labels. This approach allows each branch to explicitly avoid the interference of the excluded labels. Consequently,  instances affected by these excluded labels can benefit from the pseudo-labels provided by the corresponding branch. Note that the auxiliary model will not be involved in the testing time, which is similar to \cite{xu2021instance,qiao2023decompose}. Moreover, given mild assumptions, we demonstrate that pseudo-labels generated from the model trained in the label subspace could exhibit greater consistency with the Bayes optimal classifier compared to those from the predictive model itself. Our contributions can be summarized as follows:

\begin{itemize}[topsep=0pt,leftmargin=10pt,itemsep=0pt]    
	\item Theoretically, we prove that reduction-based pseudo-labels generated from the model trained in label subspace could be consistent with the Bayes optimal classifier than that from the predictive model itself under mild assumptions.   
    \item Practically, we propose a novel pseudo-label generation approach named {\ours} to deal with the ID-PLL problem, which utilizes the auxiliary multi-branch model with each branch trained in a subspace of labels to alleviate the impact of incorrect candidate labels strongly related to the instances. 
\end{itemize}

\section{Related Work}
PLL has been studied along two different routes: the identification-based route and the average-based route. Identification-based approaches \cite{jin2002learning,nguyen2008classification,liu2012conditional,chen2014ambiguously,yu2016maximum} are intuitional since they aim to gradually identify latent correct labels from candidate labels or eliminate incorrect labels out of candidate labels, which is also commonly referred to as disambiguating. In recent years, most researchers have attempted to tackle the problem of PLL along the identification-based route and achieved tremendous improvements. Average-based approaches \cite{hullermeier2006learning,cour2011learning,zhang2015solving} tend to deal with a learning objective where all candidate labels are treated equally. A typical average-based approach is based on distance between instances and predict the label of a unseen instance by voting among the candidate labels of its nearest neighbors in the feature space. Very recently, \cite{lv2021robustness} theoretically and empirically demonstrate their proposed approach along the average-based route have an advantage in robustness. In this paper, we still choose to follow the route of identification-based routes to handle ID-PLL.

Identification in PLL are performed by various means. In traditional PLL, to which deep neural networks (DNN) have not been applied, \cite{nguyen2008classification} maximize the margin between the maximum modeling output from candidate labels and that from non-candidate labels to implicitly identify correct labels, while \cite{yu2016maximum} further try to directly maximum the margin between the correct label and all labels. \cite{liu2012conditional} propose the Logistic Stick-Breaking Conditional Multinomial Model to maximum the likelihood of candidate label, marginalizing away the latent correct labels. \cite{zhang2016partial,feng2018leveraging,wang2019adaptive,xu2019partial} leverage the topological information in the feature space to iteratively update the confidence of each candidate label or the label distribution, from which we could determine correct and incorrect labels. In deep PLL, \cite{yao2020deep} enhance the identification ability of the model by designing an entropy-based regularization term and temporally assembling predictions of the model as the guidance of training. \cite{lv2020progressive} propose a progressive identification method that normalizes the output of the model on candidate labels in each epoch as the weight in the cross-entropy loss. \cite{yao2020network} reduce identification error by introducing a network-cooperation mechanism. \cite{feng2020provably} build a risk-consistent estimator and a classifier-consistent estimator relying on the process of identidication. \cite{wen2021leveraged} propose a loss function with a parameter weighting losses on candidate and non-candidate labels, implicitly enhancing idetification. \cite{zhang2021exploiting}, \cite{he2022partial} and \cite{lyu2022deep} improve identification through class activation value, semantic label representation and structured representation provided by deep graph neural networks, respectively. \cite{wang2022pico} propose a contrastive learning framework that could adapt to PLL and leverage the class prototypes in the framework for identification. \cite{wu2022revisiting} augment the data to obtain conformed label distributions capable of identification to perform manifold regularization. However, these works have been actually considering PLL by unrealistically assuming that incorrect candidate labels are uniformly sampled. 

From \cite{xu2021instance}, identification in ID-PLL, which is more practical, begins to be noticed. \cite{xu2021instance} explicitly estimate the label distribution for each instance, which reflect the possibility that a label is selected into candidate labels, through variational inference. \cite{xia2022ambiguity} induce a contrastive learning framework from ambiguity and obtain more identifiable representation in ID-PLL. \cite{qiao2023decompose} model the generation process of instance-dependent candidate labels to perform maximum-a-posterior, implicitly identifying the latent correct labels with prior information. \cite{xu2023progressive} propose a theoretically-guaranteed method leveraging the margin between the model output values on candidate labels to progressively identify incorrect candidate labels. \cite{he2023candidate} perform identification with selection of well-disambiguated samples. \cite{wu2024distilling} build a self-distillation framework rectifying the identification results of the teacher model to enhance the reliability of distilled knowledge. In this paper, we propose to explicitly alleviate the influence of incorrect candidate labels on the models through pseudo-labels, further enhancing identification.

\section{Proposed Method}

\subsection{Preliminaries}

First of all, we briefly introduce some necessary notations. Let $\mathcal{X}=\mathbb{R}^q$ be the $q$-dimensional instance space and $\mathcal{Y}=\{1, 2, ..., c\}$ be the label space with $c$ class labels. Given the PLL training dataset $\mathcal{D}=\{(\bm{x}_i, S_i) | 1\leq i \leq n\}$ where $\bm{x}_i \in  \mathcal{X}$ denotes the $q$-dimensional instance and $S_i \subset \mathcal{Y}$ denotes the candidate label set associated with $\bm{x}_i$. In PLL, the correct label $ y_{\bm{x}_i} $ of the instance $\bm{x}_i$ must be in the candidate label set, i.e., $ y_{\bm{x}_i} \in S_i $, and each
candidate label set $ S_i$ should not be the empty set nor the whole label set, i.e., $ S_i \notin \{\emptyset, \mathcal{Y}\}$. Besides, we do not consider the case that the candidate label set $S_i$ only has the correct label $y_{\bm{x}_i}$ in this paper, namely, $S_i \neq \{ {y_{\bm{x}_i}} \}$, For each candidate label set  $S_i$ in the training dataset, we also use the logical label vector $\bm{l}_i=[l_i^1, l_i^2, ..., l_i^c]^{\top}\in \{0,1\}^c$ to represent whether the label $j$ is one of the annotated labels, i.e., $l_i^j = 1$ if $j\in S_i$, otherwise $l_i^j = 0$. 

Let the posterior probability vector $\eta(\bm{x}) = [\eta^1(\bm{x}), \eta^2(\bm{x}), \dots, \eta^c(\bm{x})]$ with $\eta^j=p(y=j | \bm{x})$ denoting the posterior probability of the label $j$ given the instance $\bm{x}$. A Bayes optimal classifier $\eta^{\star}$ can be calculated using $\eta$, i.e.,  
\begin{equation}
    \eta^{\star}(\bm{x}) = \arg \max_{j\in \mathcal{Y}} \eta^j(\bm{x})  
\end{equation}

Moreover, let $\bar{\mathcal{Y}}$ be the labels excluded from the label space $\mathcal{Y}$ and $\eta'(\bm{x}) = [\eta'^1(\bm{x}), \eta'^2(\bm{x}), \dots, \eta'^c(\bm{x})]$ be the posterior probability without $\bar{\mathcal{Y}}$. The $j$-th element of $\eta'(\bm{x})$ denotes the posterior probability of the label $j$ given the instance $\bm{x}$ and $y \notin \bar{\mathcal{Y}}$, i.e.,
\begin{equation}\label{eta'}
    \eta'^{j}(\bm{x}) = p(y=j | \bm{x}, y \notin \bar{\mathcal{Y}}) = \left\{
    \begin{aligned}
         &\frac{\eta^{j}(\bm{x})}{1 - \sum_{k\in \bar{\mathcal{Y}}}\eta^{k}(\bm{x})}, &\text{if} \, j\notin \bar{\mathcal{Y}} \\
         & 0, \, &\text{otherwise}.
    \end{aligned}
    \right.
\end{equation}

We consider the task of PLL is to learn a score function $f: \mathcal{X} \mapsto \Delta^{c-1}$ , where $\Delta^{c-1}$ denotes the $c$-dimension simplex, as our classifier with its prediction $h(\bm{x})=\arg \max_{j\in \mathcal{Y}} f_j(\bm{x})$ consistent with that of the Bayes optimal classifier $\eta^{\star}(\bm{x})$. During the training process, the classifier $f$ could be considered to be optimized by the following objective:
\begin{equation}\label{wce}
    \mathcal{L}(f(\mathbf{X};\bm{\Theta}), \mathbf{Q}) = - \frac{1}{n} \sum_{i=1}^{n}\ell (f^{j}(\bm{x}_i; \mathbf{\Theta}), \bm{q}_i), 
\end{equation}
where the classifier $f$ is parameterized by $\bm{\Theta}$, $\ell$ denotes the cross-entropy function, $\mathbf{Q}=[\bm{q}_1, \bm{q}_2, \dots, \bm{q}_n]^{\top}$ is the pseudo-label matrix with each element $\bm{q}_i = [q_i^1, q_i^2, \dots, q_i^c]$ denoting the pseudo-label of the instance $\bm{x}_i$, satisfying $\sum_{j=1}^{c}q_i^j = 1$ and $q_i^j=0$ if $j \notin S_i$. Since our target is to make the prediction of the classifier $f$ consistent with that of the Bayes optimal classifier $\eta^{\star}$, the pseudo-label $\bm{q}_i$ should put the most mass on the label predicted by the Bayes optimal classifier $\eta^{\star}$, i.e., $\arg\max_{j\in \mathcal{Y}} q_i^j = \eta^{\star}(\bm{x})$, which is a very challenging task under instance-dependent PLL.

\subsection{Overview}

To begin with, we provide a formal definition of disturbing incorrect labels, which are labels that the predictive model finds challenging to identify as incorrect. We prove that a model trained in a label subspace excluding these disturbing incorrect labels can produce pseudo-labels more consistent with the Bayes optimal classifier for instances whose candidate label sets include these disturbing incorrect labels. Moreover, we establish a boundary for the conditional probability that the pseudo-labels of these instances are consistent with the Bayes optimal classifier.

Motivated by these theoretical results, we propose an approach, named {\ours}, which leverages reduction-based pseudo-labels to train our predictive model. The reduction-based pseudo-labels are aggregated from the output of an auxiliary model with multiple branches, each trained in a label subspace that explicitly excludes certain labels. To obtain the reduction-based pseudo-label, we use a meta-learned weight vector to combine the outputs of all branches. This method mitigates the influence of incorrect candidate labels and enhances the performance of the predictive model.

\subsection{The proposed approach}

We first introduce labels which are easy to be carelessly selected as candidate labels during annotation and hard to be distinguished as incorrect labels according to the output of the predictive model during training. These labels pose great challenge to generate pseudo-labels $\bm{q}_i$ consistent with the Bayes optimal classifier $\eta^{\star}(\bm{x}_i)$.

\begin{definition}\label{def}
({$(\tau, f, \epsilon)$}-disturbing incorrect label) An incorrect label $j$ is said to be $(\tau, f, \epsilon)$-disturbing for the predictor $f$ on some instance $\bm{x}$ with $\eta^{\star}(\bm{x}) \neq j$ if $\exists \epsilon \in (0, 1), \, \forall j \in \mathcal{Y}, \, \max_{\bm{x}} \vert f^j(\bm{x}) - \eta^j(\bm{x}) \vert \leq \epsilon $ and $\exists \tau \in (0, \min\{1, 2\epsilon\}],$ the posterior $ \eta^{\eta^{\star}(\bm{x})}(\bm{x}) - \eta^j(\bm{x}) \leq \tau$.
\end{definition}

Here, $\tau$ indicates the degree that the posterior $\eta^j(\bm{x})$ approaches $ \eta^{\eta^{\star}(\bm{x})}(\bm{x})$. The smaller its value, the easier the label $j$ is selected into the candidate label set. $\epsilon$ indicates the degree that the predictive model $f$ approximates the posterior $\eta$. According to \cite{xu2023progressive}, if $ \eta^{\eta^{\star}(\bm{x})}(\bm{x}) - \eta^j(\bm{x}) > 2\epsilon$, we could distinguish label $j$ as incorrect labels according to the output of the predictive model.

Then we analyze the pseudo-labels of the instances, whose indistinguishable labels is denoted by $\bar{\mathcal{Y}}$. An auxiliary model $\varphi$ is considered to train without the influence of indistinguishable labels. On mild assumptions, we prove that the pseudo-label provided by the auxiliary model $\varphi$ has more chance to be consistent with the Bayes optimal classifier than the predictive model. The proof can be found in Appendix \ref{proof:theorem1}.
\begin{theorem}\label{theorem1}
    Let $J(\bm{x}, \bar{\mathcal{Y}}) = \{ \bm{x} | \forall j \in \bar{\mathcal{Y}}, j \, \text{is a} \, (\tau, f, \epsilon)\text{-disturbing incorrect label for } \, \bm{x} \, \text{with} \, y \neq j, \, \text{and} \, \forall k\notin \{y\} \cup \bar{\mathcal{Y}}, \eta^{k}(\bm{x}) < \eta^{j}(\bm{x}) \}$. Suppose that a model $\varphi$ trained without the label space $\bar{\mathcal{Y}}$ satisfies $\exists \epsilon' \in (0, \min \{1, \min_{\bm{x}\in \mathcal{J}(\bm{x}, \bar{\mathcal{Y}})} \frac{(\eta^{\eta^{\star}(\bm{x})}(\bm{x}) - \eta^b(\bm{x}))(\eta^{\eta^{\star}(\bm{x})}(\bm{x}) -\eta^a(\bm{x}))}{4\epsilon(1 - \sum_{j \in \bar{\mathcal{Y}}}\eta^j(\bm{x}))}\})$ with $a = \arg\max_{j\in \bar{\mathcal{Y}}}\eta^{j}(\bm{x})$ and $b = \arg\max_{j\notin \{y\}\cup\bar{\mathcal{Y}}}\eta^{j}(\bm{x})$, $\vert \varphi^{j}(\bm{x}) - \eta'^{j}(\bm{x}) \vert \leq \epsilon'$, we could obtain:
    \begin{equation}
        p(\eta^{\star}(\bm{x}) = \arg\max_{j\in \mathcal{Y}} q^j | \bm{x} \in J(\bm{x}, \bar{\mathcal{Y}})) \leq p(\eta^{\star}(\bm{x}) = \arg\max_{j\in \mathcal{Y}} q'^j | \bm{x} \in J(\bm{x}, \bar{\mathcal{Y}}))
    \end{equation}
    
\end{theorem}

Additionally, we further analyze the chance that the pseudo-label provided by the auxiliary model $\varphi$ is consistent with the Bayes optimal classifier. We assume the Tsybakov condition \cite{tsybakov2004optimal,zheng2020error} holds around the margin of the decision boundary of the true posterior in the multi-class scenario. 
\begin{assumption} \label{Tsybakov}
    (multi-class Tsybakov condition) $\exists C, \lambda > 0$ and $\exists t_0 \in (0, 1]$, such that for all $t \leq t_0$, 
    \begin{equation}
        p(\eta^{\eta^{\star}(\bm{x})}(\bm{x}) - \eta^{s(\bm{x})}(\bm{x}) \leq t) \leq Ct^{\lambda},
    \end{equation}
    where $s(\bm{x}) = \arg\max_{j\in \mathcal{Y}, j\neq \eta^{\star}(\bm{x})} \eta^j(\bm{x})$ denotes the second best prediction of $\eta(\bm{x})$.
\end{assumption}

Under Assumption \ref{Tsybakov}, we could prove the pseudo-label provided by the auxiliary model $\varphi$ has a good chance to be consistent with the Bayes optimal classifier. The proof can be found in Appendix \ref{proof:theorem2}.
\begin{theorem}\label{theorem2}
    Suppose that for $\bm{x} \in J(\bm{x}, \bar{\mathcal{Y}})$, its posterior $\eta(\bm{x})$ fulfills Assumption \ref{Tsybakov} for constants $C, \lambda > 0$ and $t_0 \in (0, 1]$. Suppose that a model $\varphi$ trained without the label space $\bar{\mathcal{Y}}$ satisfies $\exists \epsilon' \in (0, \min \{1, \min_{\bm{x}\in \mathcal{J}(\bm{x}, \bar{\mathcal{Y}})} \frac{(\eta^{\eta^{\star}(\bm{x})}(\bm{x}) - \eta^b(\bm{x}))(\eta^{\eta^{\star}(\bm{x})}(\bm{x}) -\eta^a(\bm{x}))}{4\epsilon(1 - \sum_{j \in \bar{\mathcal{Y}}}\eta^j(\bm{x}))}\})$ with $a = \arg\max_{j\in \bar{\mathcal{Y}}}\eta^{j}(\bm{x})$ and $b = \arg\max_{j\notin \{y\}\cup\bar{\mathcal{Y}}}\eta^{j}(\bm{x})$, $\vert \varphi^{j}(\bm{x}) - \eta'^{j}(\bm{x}) \vert \leq \epsilon'$, we could obtain:
    \begin{equation}
        p(\eta^{\star}(\bm{x}) = \arg\max_{j\in \mathcal{Y}} q'^j | \bm{x} \in J(\bm{x}, \bar{\mathcal{Y}})) \geq 1 - C[O(\epsilon\epsilon')]^{\lambda}
    \end{equation}
\end{theorem}

\begin{algorithm}[t]
	\caption{{\ours} Algorithm}
	\label{alg:algorithm}
	\textbf{Input}: PLL training dataset $\mathcal{D} = \{(\bm{x}_{i}, S_{i}| 1 \leq i \leq n\}$, validating dataset $\mathcal{D}^{\text{val}}=\{ (\bm{x}_{i}^{\text{val}}, \bm{y}_i^{\text{val}} | 1 \leq i \leq n^{\text{val}} \}$, Epoch $I$, Iteration $K$; \\
	\textbf{Output}: The predictive model $f(\cdot; \bm{\Theta})$
	\begin{algorithmic}[1] 
		\STATE Initialize the parameters of the predictive model, the auxiliary model and meta-learner, i.e., $\bm{\Theta}^{0}$, $\{\bm{\Omega}_j^{0}\}_{j=1}^{c}$ and $\bm{\Gamma}^{0}$;
		\FOR{$i=1,2,\dots,I$}
		\STATE Randomly shuffle the training dataset $\mathcal{D}$ and divide it into $K$ mini-batches;
		\FOR{$k=0,1,\dots,K-1$}
		
		\STATE Calculate $\mathbf{U}_i$ for the instance $\bm{x}_i$ according to Eq. (\ref{eq:U});
		\STATE Update $\mathbf{\Omega}_j^k$ to $\mathbf{\Omega}_j^{k+1}$ according to Eq. (\ref{eq:update_Omega});
        \STATE Calculate $\bm{v}_i$ for the instance $\bm{x}_i$ according to Eq. (\ref{eq:v});
        \STATE Save $\mathbf{\Theta}_j^{k}$ and update $\mathbf{\Theta}_j^{k}$ to $\mathbf{\Theta}_j^{k+1}$ according to Eq. (\ref{eq:update_Theta_1});
        \STATE Randomly sample a mini-batch from $\mathcal{D^{\text{val}}}$ and update $\mathbf{\Gamma}_j^{k}$ to $\mathbf{\Gamma}_j^{k+1}$ according to Eq. (\ref{eq:update_Gamma});
        \STATE Calculate $\bm{q}_i$ for the instance $\bm{x}_i$ according to Eq. Eq. (\ref{eq:w})(\ref{eq:v})(\ref{eq:q});
        \STATE Rollback to $\mathbf{\Theta}_j^{k}$ and update $\mathbf{\Theta}_j^{k}$ to $\mathbf{\Theta}_j^{k+1}$ according to Eq. (\ref{eq:update_Theta_2});
		\ENDFOR
		\ENDFOR
	\end{algorithmic}
\end{algorithm}

Our theoretical insight inspires a new algorithm for the generation of the pseudo-label $\bm{q}_i$ in the optimization objective Eq. (\ref{wce}). To begin with, we decompose the pseudo-label $\bm{q}_i$ into the basic pseudo-label $\bm{\mu}_i$ and the reduction-based pseudo-label $\bm{v}_i$, i.e., 
\begin{equation}\label{eq:q}
    \bm{q}_i = \alpha \bm{\mu}_i + (1 - \alpha) \bm{v}_i,
\end{equation}
where $\alpha$ is a trade-off hyper-parameter to decide the influence of the introduced reduction-based pseudo-labels. The basic pseudo-label $\bm{\mu}_i$ is initialized with uniform weights on candidate labels and then could be calculated by using the outputs of the predictive model $f$:
\begin{equation}\label{eq:mu}
    \mu_i^{j} = \left\{
    \begin{aligned}
        & \frac{f^j(\bm{x}_i; \mathbf{\Theta})}{\sum_{k\in S_i} f^k(\bm{x}_i; \mathbf{\Theta})} \, &\text{if} \, j\in S_i, \\
        & 0, \, &\text{otherwise}, \\
    \end{aligned} 
    \right.
\end{equation}
which puts more weights on more possible candidate labels \cite{lv2020progressive}. The reduction-based pseudo-label $\bm{v}_i$ can be obtained by the output of the model trained without the influence of the label from $1$ to $j$ formulated as $\mathbf{U}_i$ and a vector $\bm{w}_i$ to weight these output:
\begin{equation}\label{eq:v}
    \bm{v}_i = \bm{w}_i \mathbf{U}_i.
\end{equation}

Here, $\mathbf{U}_i = [\bm{u}_i^1, \bm{u}_i^2, \dots, \bm{u}_i^c]^{\top} \in \mathbb{R}^{c\times c}$ is a reduction-based matrix, the $j$-th row of which is initialized with uniform weights on candidate labels without label $j$ and then calculated by:
\begin{equation}\label{eq:U}
    u_i^{j,r} = \left\{
    \begin{aligned}
        & \frac{\varphi^{r}(\bm{z}_i; \bm{\Omega}_j)}{\sum_{k\in S_i \setminus \{j\}} \varphi^{k}(\bm{z}_i; \bm{\Omega}_j)} \, &\text{if} \, r\in S_i \setminus \{j\}, \\
        & 0, \, &\text{otherwise}, \\
    \end{aligned} 
    \right.
\end{equation}
where $\varphi$ is an auxiliary model parameterized by $\{\mathbf{\Omega}_j\}_{j=1}^{c}$ to form $c$ branches, and $\bm{z}_i$ is the extracted features given the instance $\bm{x}_i$. The $j$-th branch $\varphi(\cdot; \mathbf{\Omega}_j)$ is trained without $j$ in the label space $\mathcal{Y}$, and the loss function for the auxiliary model $\varphi$ could be formulated by: 
\begin{equation}\label{loss:branch}
    \mathcal{L}^{\text{aux}}(\{\varphi(\bm{Z}; \bm{\Omega}_j)\}_{j=1}^{c}, \{\mathbf{U}_i\}_{i=1}^{n}) = - \frac{1}{n}\sum_{i=1}^{n}\sum_{j=1}^{c} \ell( \varphi(\bm{z}_i; \bm{\Omega}_j), \bm{u}_i^j ).
\end{equation}
In this way, the $j$-th branch $\varphi(\cdot; \mathbf{\Omega}_j)$ could be considered as an approximation of $\eta'(\bm{x})$ with $\bar{\mathcal{Y}}=\{j\}$.

And $\bm{w}_i = [w_i^1, w_i^2, \dots, w_i^c] \in \mathbb{R}^{1\times c}$ is a weight vector output by a model $g$ parameterized by $\mathbf{\Gamma}$ given the instance $\bm{x}_i$, i.e., 
\begin{equation}\label{eq:w}
\bm{w}_i = g(\bm{x}_i; \mathbf{\Gamma}).
\end{equation}
Since it is unknown which label in the candidate set $S_i$ of the instance $\bm{x}_i$ is the label interfering the correct label $y_{\bm{x}_i}$, we formulate the model $g$ as a meta-learner and learn-to-learn a weight vector to eliminate the disturbance of incorrect candidate labels and obtain the reduction-based pseudo-label $\bm{v}_i$ for training. We employ the reduction-based pseudo-labels $\mathbf{V}=[\bm{v}_1, \bm{v}_2, \dots, \bm{v}_n]$ with each element $\bm{v}_i$ calculated from $\bm{w}_i$ to update the predictive model $f(\cdot; \mathbf{\Theta})$: 
\begin{equation}\label{loss:inner}
    \mathcal{L}^{\text{inner}}(f(\mathbf{X};\bm{\Theta}), \mathbf{V
    }) = - \frac{1}{n}\sum_{i=1}^{n}\ell(f(\bm{x}_i; \bm{\Theta}), \bm{v}_i),
\end{equation}

Then, we assess the updated predictive model on the validation dataset $\mathcal{D}^{\text{val}}=\{ (\bm{x}_{i}^{\text{val}}, \bm{y}_i^{\text{val}} | 1 \leq i \leq n^{\text{val}} \}$ to update the meta-learner $g(\cdot; \mathbf{\Gamma})$
\begin{equation}\label{loss:outer}
    \mathcal{L}^{\text{outer}}(f(\mathbf{X}^{\text{val}};\bm{\Theta}), \mathbf{Y}^{\text{val}}) = -\frac{1}{n^{\text{val}}} \sum_{i=1}^{n^{\text{val}}} \ell(f(\bm{x}_i^{\text{val}}; \bm{\Theta}), \bm{y}_i^{\text{val}}).
\end{equation}

Overall, the meta-learning objective can be formulated as a bi-level optimization problem as follows:
\begin{equation}\label{loss:meta-objective}
\begin{aligned}
    &\bm{\Gamma}^{\star} = \arg\min_{\bm{\Gamma}} \mathcal{L}^{\text{outer}}(f(\mathbf{X}^{\text{val}};\bm{\Theta}^{\star}(\bm{\Gamma})), \mathbf{Y}^{\text{val}}) \\
    \mathrm{ s.t. } \quad &\bm{\Theta}^{\star} = \arg\min_{\bm{\Theta}} \mathcal{L}^{\text{inner}}(f(\mathbf{X};\bm{\Theta}), \mathbf{V}),
\end{aligned}    
\end{equation}

To solve the optimization of Eq. (\ref{loss:meta-objective}), an online strategy inspired by \cite{shu2029meta} is employed to update $\mathbf{\Theta}$ and $\mathbf{\Gamma}$ through a single optimization loop, respectively, which guarantees the efficiency of the algorithm. Specifically, we shuffle the training set $\mathcal{D}$ into $K$ mini-batches. Each mini-batch contains $m$ examples, i.e., $\{(\bm{x}_i, S_i) | 1\leq i \leq m\}$. In the step $k$ of training, we employ stochastic gradient descent (SGD) to optimize the meta-learning objective $\mathcal{L}^{\text{inner}}$ and $\mathcal{L}^{\text{outer}}$ with the loss functions for the classifier and auxiliary model $\mathcal{L}$ and $\mathcal{L}^{\text{aux}}$ on the $k$-th mini-batch. 

\begin{table*}[t]
	\centering
	\renewcommand{\arraystretch}{1.1}
\setlength{\tabcolsep}{10pt}
	\caption{ Classification accuracy (mean$\pm$std) of each comparing approach on  benchmark datasets for instance-dependent PLL }
	\begin{tabular}{cccccc}
        \toprule
		Datasets & CIFAR10 & CIFAR100 & TinyImageNet \\ 
        \midrule
        {\ours} & \textbf{87.53$\pm$0.21\%} & \textbf{65.03$\pm$0.21\%} & \textbf{40.74$\pm$0.64\%} \\ 
        \midrule
        {\pldirk} & 84.63$\pm$0.22\%$\bullet$ & 58.17$\pm$0.20\%$\bullet$ & 25.77$\pm$1.55\%$\bullet$ \\ 
        {\plsdct} & 86.50$\pm$0.13\%$\bullet$ & 60.95$\pm$0.35\%$\bullet$ & 36.50$\pm$0.35\%$\bullet$ \\ 
        {\plpop} & 86.23$\pm$0.36\%$\bullet$ & 60.71$\pm$0.16\%$\bullet$ & 39.27$\pm$0.86\%$\bullet$ \\ 
        {\plidgp} & 86.43$\pm$0.23\%$\bullet$ & 64.38$\pm$0.27\%$\bullet$ & 32.21$\pm$1.14\%$\bullet$ \\ 
        {\plable} & 85.11$\pm$0.24\%$\bullet$ & 61.21$\pm$0.37\%$\bullet$ & 23.60$\pm$0.77\%$\bullet$ \\ 
        {\vpll} & 85.48$\pm$0.62\%$\bullet$ & 62.96$\pm$0.96\%$\bullet$ & 37.14$\pm$0.21\%$\bullet$ \\ 
        \midrule
        {\plcr} & 86.37$\pm$0.38\%$\bullet$ & 64.12$\pm$0.23\%$\bullet$ & 24.59$\pm$1.68\%$\bullet$ \\ 
        {\pico} & 86.16$\pm$0.21\%$\bullet$ & 62.98$\pm$0.38\%$\bullet$ & 29.95$\pm$0.48\%$\bullet$ \\ 
        {\cavl} & 59.67$\pm$3.30\%$\bullet$ & 52.59$\pm$1.01\%$\bullet$ & 28.10$\pm$0.77\%$\bullet$ \\ 
        {\lws} & 37.49$\pm$2.82\%$\bullet$ & 53.98$\pm$0.99\%$\bullet$ & 27.37$\pm$0.82\%$\bullet$ \\ 
        {\rcpll} & 85.95$\pm$0.40\%$\bullet$ & 63.41$\pm$0.56\%$\bullet$ & 35.74$\pm$0.61\%$\bullet$ \\ 
        {\ccpll} & 79.96$\pm$0.99\%$\bullet$ & 62.40$\pm$0.84\%$\bullet$ & 31.46$\pm$1.24\%$\bullet$ \\ 
        {\proden} & 86.04$\pm$0.21\%$\bullet$ & 62.56$\pm$1.49\%$\bullet$ & 33.37$\pm$0.97\%$\bullet$ \\ 
        \bottomrule
	\end{tabular}
	\label{benchmark}
\end{table*}

First, as for the auxiliary model $\varphi$, we update the parameter of each branch $\mathbf{\Omega}_j^{k}$ to $\mathbf{\Omega}_j^{k+1}$ as follows: 
\begin{equation}\label{eq:update_Omega}
    \mathbf{\Omega}_j^{k+1} = \mathbf{\Omega}_j^{k} - \frac{\beta_1}{m}\sum_{i=1}^{m}\frac{\partial \ell( \varphi(\bm{z}_i; \bm{\Omega}_j^{k}), \bm{u}_i^j )}{\partial \mathbf{\Omega}_j^{k}}, 
\end{equation}
where $\beta_1$ is the step size. Then, after updating $\mathbf{\Omega}_j^{k}$ to $\mathbf{\Omega}_j^{k+1} $, we could obtain the reduction-based pseudo-label $\bm{v}_i$ by Eq. (\ref{eq:v}) and optimize the inner objective of the bi-level optimization Eq. (\ref{loss:meta-objective}):  
\begin{equation}\label{eq:update_Theta_1}
    \mathbf{\Theta}^{k+1} = \mathbf{\Theta}^{k} - \frac{\beta_2}{m}\sum_{i=1}^{m}\frac{\partial \ell(f(\bm{x}_i; \bm{\Theta}^{k}), \bm{v}_i)}{\partial \mathbf{\Theta}^{k}},
\end{equation}
where $\beta_2$ is the step size. Note that after updating $\mathbf{\Theta}^{k}$ to $\mathbf{\Theta}^{k+1}$ with the reduction-based pseudo-label $\bm{v}_i$, $\mathbf{\Theta}^{k+1}$ is dependent on the parameters $\mathbf{\Gamma}$ of the meta-learner $g$, i.e., $\mathbf{\Theta}^{k+1}(\mathbf{\Gamma}^{k})$, which allows the updation of $\mathbf{\Gamma}^{k}$ through the loss function $\mathcal{L}^{\text{outer}}$ as follows:
\begin{equation}\label{eq:update_Gamma}
    \mathbf{\Gamma}^{k+1} = \mathbf{\Gamma}^{k} - \frac{\beta_3}{m}\sum_{i=1}^{m}\frac{\partial \ell(f(\bm{x}_i^{\text{val}}; \bm{\Theta}^{k+1}), \bm{y}_i^{\text{val}})}{\partial \mathbf{\Gamma}^{k}},
\end{equation}
where we also randomly sample $m$ examples from $\mathcal{D^{\text{val}}}$, and $\beta_3$ is the step size. Finally, we rollback the parameters of our classifier to $\mathbf{\Theta}^{k}$ and employ the pseudo-label $\bm{q}_i$ generated by Eq. (\ref{eq:w})(\ref{eq:v})(\ref{eq:q}) to optimize it with the same step size with Eq. (\ref{eq:update_Theta_1}):
\begin{equation}\label{eq:update_Theta_2}
    \mathbf{\Theta}^{k+1} = \mathbf{\Theta}^{k} - \frac{\beta_2}{m}\sum_{i=1}^{m}\frac{\partial \ell(f(\bm{x}_i; \bm{\Theta}^{k}), \bm{q}_i)}{\partial \mathbf{\Theta}^{k}}. 
\end{equation}

In this way, as the auxiliary model parameters $\{\mathbf{\Omega}_j\}_{j=1}^{c}$ and meta-learner parameters $\mathbf{\Gamma}$ are updated iteratively, the pseudo-label $\bm{q}_i$ is also refined to contribute to the optimization of the classifier $f(\cdot; \mathbf{\Theta})$ step by step.  The algorithmic description of {\ours} is presented in Algorithm \ref{alg:algorithm}.

\section{Experiments}
In this section, we validate the effectiveness of our proposed {\ours} by conducting it on manually corrupted benchmark datasets and real-world datasets and comparing its results against DNN-based PLL algorithms. Furthermore, we explore {\ours} through ablation study, sensitivity analysis, and convergence analysis. The implementation is based on Pytorch \cite{paszke2019pytorch} and carried out with NVIDIA RTX 3090.

\begin{table*}[t]
\renewcommand{\arraystretch}{1.1}
\setlength{\tabcolsep}{6pt}
	\centering
	\small
	\caption{Classification accuracy (mean$\pm$std) of comparing algorithms on  the real-world datasets.  }
	\begin{tabular}{cccccc}
		\toprule
		Datasets & Lost & BirdSong & MSRCv2 & Soccer Player & Yahoo!News \\
        \midrule
        {\ours} & \textbf{81.07$\pm$0.74\%} & \textbf{75.27$\pm$0.23\%} & \textbf{51.65$\pm$0.65\%} & \textbf{56.94$\pm$0.34\%} & \textbf{68.01$\pm$0.19\%} \\ 
        \midrule
        {\pldirk} & 79.24$\pm$0.63\%$\bullet$ & 74.52$\pm$0.23\%$\bullet$ & 48.59$\pm$0.28\%$\bullet$ & 55.83$\pm$0.35\%$\bullet$ & 67.65$\pm$0.32\%$\bullet$ \\ 
        {\plpop} & 78.57$\pm$0.45\%$\bullet$ & 74.47$\pm$0.36\%$\bullet$ & 45.86$\pm$0.28\%$\bullet$ & 54.48$\pm$0.10\%$\bullet$ & 66.38$\pm$0.07\%$\bullet$ \\ 
        {\plidgp} & 77.02$\pm$0.80\%$\bullet$ & 74.23$\pm$0.17\%$\bullet$ & 50.45$\pm$0.47\%$\bullet$ & 55.99$\pm$0.28\%$\bullet$ & 66.62$\pm$0.19\%$\bullet$ \\ 
        {\vpll} & 76.87$\pm$0.86\%$\bullet$ & 73.39$\pm$0.26\%$\bullet$ & 49.97$\pm$0.43\%$\bullet$ & 55.81$\pm$0.10\%$\bullet$ & 66.26$\pm$0.13\%$\bullet$ \\ 
        \midrule
        {\cavl} & 75.89$\pm$0.42\%$\bullet$ & 73.47$\pm$0.13\%$\bullet$ & 44.73$\pm$0.96\%$\bullet$ & 54.06$\pm$0.67\%$\bullet$ & 65.44$\pm$0.23\%$\bullet$ \\ 
        {\lws} & 73.13$\pm$0.32\%$\bullet$ & 51.45$\pm$0.26\%$\bullet$ & 49.85$\pm$0.49\%$\bullet$ & 50.24$\pm$0.45\%$\bullet$ & 48.21$\pm$0.29\%$\bullet$ \\ 
        {\rcpll} & 76.26$\pm$0.46\%$\bullet$ & 69.33$\pm$0.32\%$\bullet$ & 49.47$\pm$0.43\%$\bullet$ & 56.02$\pm$0.59\%$\bullet$ & 63.51$\pm$0.20\%$\bullet$ \\ 
        {\ccpll} & 63.54$\pm$0.25\%$\bullet$ & 69.90$\pm$0.58\%$\bullet$ & 41.50$\pm$0.44\%$\bullet$ & 49.07$\pm$0.36\%$\bullet$ & 54.86$\pm$0.48\%$\bullet$ \\ 
        {\proden} & 76.47$\pm$0.25\%$\bullet$ & 73.44$\pm$0.12\%$\bullet$ & 45.10$\pm$0.16\%$\bullet$ & 54.05$\pm$0.15\%$\bullet$ & 66.14$\pm$0.10\%$\bullet$ \\
		\bottomrule
	\end{tabular}
	\label{real-world}

\end{table*}

\subsection{Datasets}

{\ours} and compared DNN-based PLL algorithms are implemented on three widely used benchmark datasets in deep learning: \texttt{CIFAR-10}, \texttt{CIFAR-100} \cite{krizhevsky2009learning} and \texttt{TinyImageNet} \cite{le2015tiny}. For these datasets, we generate instance-dependent candidate labels through the same strategy as \cite{xu2021instance}, which considers instance-dependent PLL for the first time, to create manually corrupted benchmark datasets.

Besides, since data augmentation cannot be performed on extracted features from audio and video data, our approach and data-augmentation-free PLL methods are also performed on five frequently used real-world datasets, which come from different practical application domains, including \texttt{Lost} \cite{cour2011learning}, \texttt{BirdSong} \cite{briggs2012rank}, \texttt{MSRCv2}  \cite{liu2012conditional}, \texttt{Soccer Player} \cite{zeng2013learning} and \texttt{Yahoo!News} \cite{guillaumin2010multiple}. 

As for benchmark datasets, we split $10\%$ samples from the training datasets to form the validating datasets. As for real-world datasets, we conduct the algorithms with $80\%$/$10\%$/$10\%$ train/validation/test split. Then we run five trials on each datasets with different random seeds and report the mean accuracy and standard deviation of all comparing algorithms.

\begin{table}[t]
	\centering
\renewcommand{\arraystretch}{1.1}
\setlength{\tabcolsep}{10pt}
	\caption{Classification accuracy (mean$\pm$std) for comparison against {\oursnm}.}
	\label{Ablation}
	\begin{tabular}{lcc}
		\toprule
		Dataset & {\ours} & {\oursnm} \\ 
        \midrule
        CIFAR-10 & \textbf{87.53$\pm$0.21\%} & 85.43$\pm$0.52\%$\bullet$ \\ 
        CIFAR-100 & \textbf{65.03$\pm$0.21\%} & 61.18$\pm$0.34\%$\bullet$ \\ 
        TinyImageNet & \textbf{40.74$\pm$0.64\%} & 35.35$\pm$1.01\%$\bullet$ \\
        \midrule
        Lost & \textbf{81.07$\pm$0.74\%} & 77.14$\pm$1.62\%$\bullet$ \\ 
        BirdSong & \textbf{75.27$\pm$0.23\%} & 73.18$\pm$0.71\%$\bullet$ \\ 
        MSRCv2 & \textbf{51.65$\pm$0.65\%} & 46.62$\pm$1.54\%$\bullet$ \\ 
        Soccer Player & \textbf{56.94$\pm$0.34\%} & 55.48$\pm$0.65\%$\bullet$ \\ 
        Yahoo!News & \textbf{68.01$\pm$0.19\%} & 66.82$\pm$0.13\%$\bullet$ \\ 
		\bottomrule
	\end{tabular}
\end{table}

\subsection{Baselines}

We compare {\ours} with six methods, which are designed for the challenge of ID-PLL:
\begin{itemize}[topsep=0pt,leftmargin=10pt,itemsep=0pt] 
    \item  {\pldirk} \cite{wu2022revisiting}, a self-distillation framework which rectifies the label confidences as the distilled knowledge to guide the training of the predictive model. 
    \item  {\plsdct} \cite{wang2022pico}, a sample selection framework which selects well-disambiguated samples based on normalized entropy in two stages for the training of the predictive model with data augmentation.
    \item  {\plpop} \cite{xu2021instance}, a label purification framework which progressively purifies each candidate labels as the performance of the trained predictive model improves.  
    \item  {\plidgp} \cite{wu2022revisiting}, a maximum-a-posterior approach which decomposes candidate labels into the results sampled from two different distributions to form a optimization objective for training.
    \item  {\plable} \cite{wang2022pico}, a contrastive learning framework which is based on data augmentation and pulls ambiguity-induced positives closer and the remaining instances further in the representation space.
    \item  {\vpll} \cite{xu2021instance}, an encoder-decoder framework which leverages variational inference to recover latent label distributions for the guidance of training the predictive model. 
\end{itemize}
Besides, we also compare our method with another seven classical DNN-based PLL methods: 
\begin{itemize}[topsep=0pt,leftmargin=10pt,itemsep=0pt]
    \item  {\plcr} \cite{wu2022revisiting}, a manifold regularization approach which is based on data augmentation and proposes a consistency regularization objective to preserve manifold structure in feature and label space.
    \item  {\pico} \cite{wang2022pico}, a contrastive learning framework that relies on data augmentation and achieves label disambiguation through contrastive prototypes.
    \item  {\cavl} \cite{zhang2021exploiting}, a identification-based approach which a discriminative approach which identifies correct labels from candidate labels by class activation value.  
    \item  {\lws} \cite{wen2021leveraged}, an identification-based approach which introduces a leverage parameter to account for the trade-off between losses on candidate and non-candidate labels.
    \item  {\rcpll}  \cite{feng2020provably}, a risk-consistent approach utilizes the loss correction strategy to estimate the true risk by only using data with candidate labels.
    \item  {\ccpll} \cite{feng2020provably}, a classifier-consistent approach which leverages the transition matrix to learn a predictive model that could approximate the optimal one.
    \item  {\proden} \cite{lv2020progressive}, a self-training algorithm which normalizes the output of the predictive model on candidate labels and utilizes it as a weight on the cross-entropy function for training.
\end{itemize}
To ensure fairness, we utilize the same network backbone, optimizer, and data augmentation strategy across all compared methods. We take the same backbone as \cite{xu2023progressive,wu2024distilling} on \texttt{CIFAR-10}, \texttt{CIFAR-100} and all realworld datasets, and \cite{liang2022efficient,yin2023squeeze} on \texttt{TinyImageNet}. The optimizer is stochastic gradient descent (SGD) \cite{robbins1951stochastic} with momentum $0.9$, batch size $256$, and epoch $250$. We follow \cite{wu2022revisiting} to apply the data augmentation strategy. Besides, for hyper-parameters, we carefully select the most appropriate ones for each algorithm to ensure optimal model parameters based on their performances on the validation datasets. To mitigate overfitting, the training procedure of a model will be halted prematurely if its performance on the validation dataset fails to improve over 50 epochs.

\begin{figure*}[t]
	\centering
	\small
	\subfigure[Sensitivity]{
		\includegraphics[scale=0.25]{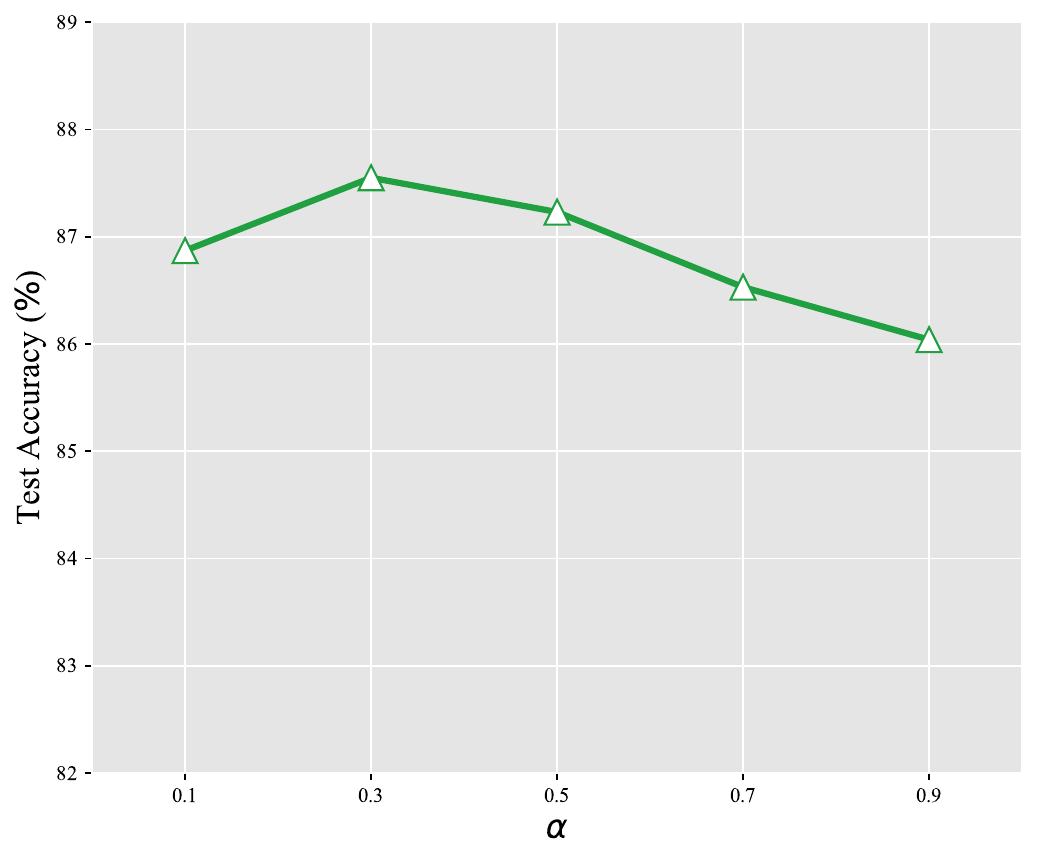}\label{sensitivity}
	}
	\hfill
	\subfigure[Consistency]{
		\includegraphics[scale=0.25]{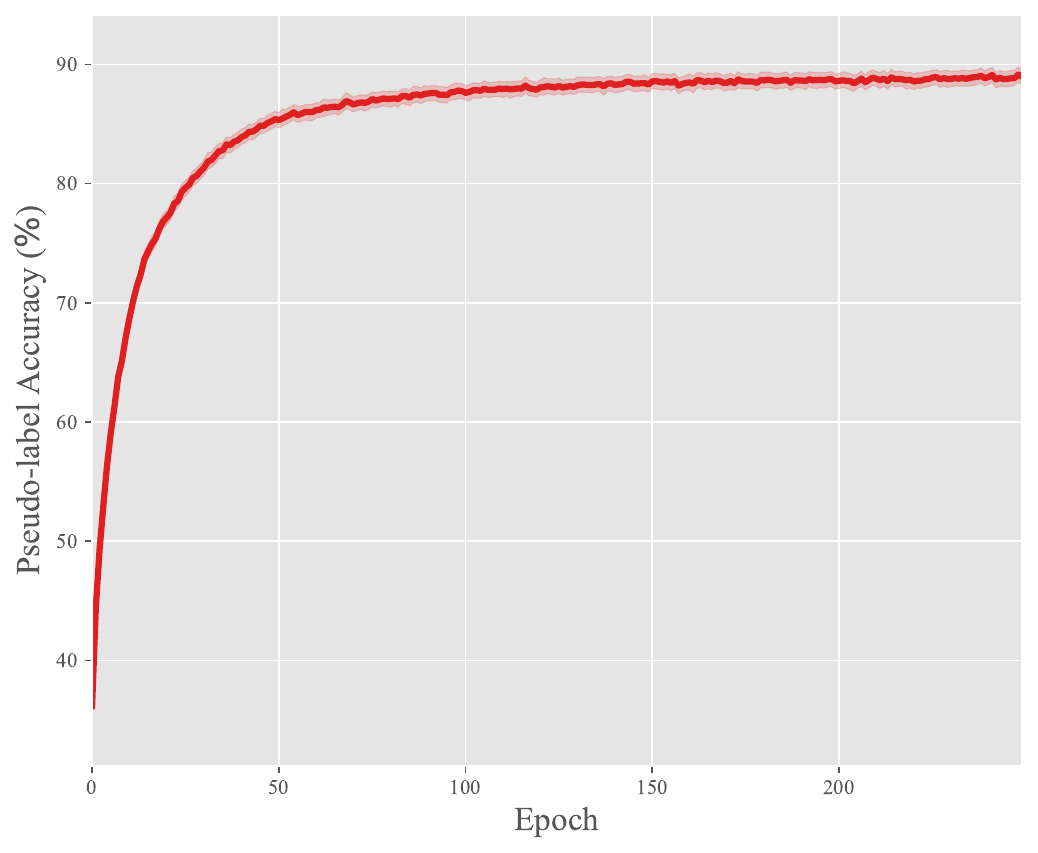}\label{consistency}
	}
	\hfill
	\subfigure[Convergence]{
		\includegraphics[scale=0.25]{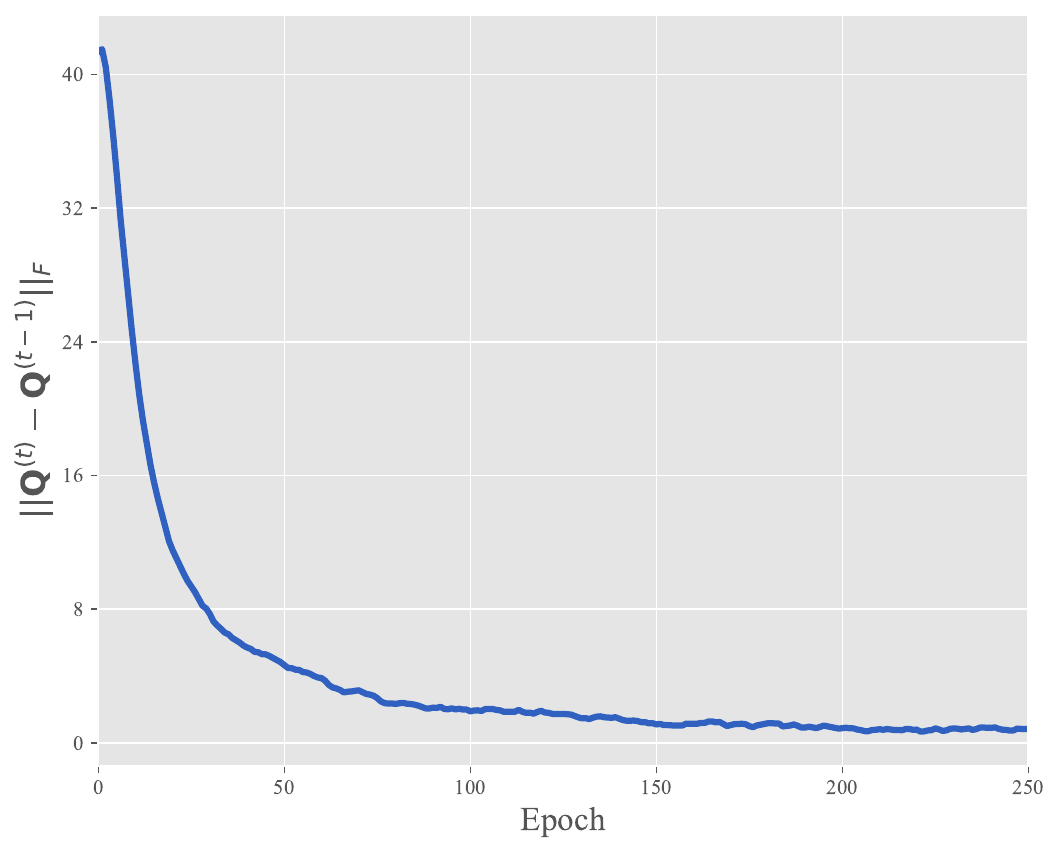}\label{convergence}
	}
	\hfill
	\caption{Further analysis of {\ours} on \texttt{CIFAR-10}}
\end{figure*}

\subsection{Experimental Results}

The performance of each DNN-based method on each corrupted benchmark dataset is summarized in Table \ref{benchmark}, where the best results are highlighted in bold and $\bullet$/$\circ$ indicates whether {\ours} statistically wins/loses to the comparing method on each dataset additionally (pairwise t-test at 0.05 significance level). Overall, we observe that {\ours} significantly outperforms all comparative methods, whether ID-PLL or classic DNN-based PLL approaches, across all benchmark datasets. The improvements are especially pronounced on the complex \texttt{TinyImageNet} dataset.

Table \ref{real-world} demonstrates the ability of {\ours} to solve the ID-PLL problem in real-world datasets. Note that data-augmentation-based algorithms, including {\plsdct}, {\plable}, {\plcr} and {\pico}, are not compared on the real-world PLL datasets due to the inability of data augmentation to be employed on the extracted features from various domains. We can find that our method still has significantly stronger competence than others on all datasets, even the large dataset \texttt{Soccer Player} and \texttt{Yahoo!News}, against all other comparing algorithms. 

\subsection{Further Analysis}

To demonstrate the effectiveness of the meta-learned weight $\bm{w}$ introduced by {\ours}, we explore a vanilla variant, {\oursnm}, where a uniform weight is applied to the candidate labels of the instance instead of the weight output by a parameterized model learned through meta-learning. The performance of {\ours} compared to {\oursnm} is assessed using classification accuracy, with pairwise t-tests conducted at a significance level of 0.05. As shown in Table \ref{Ablation}, by leveraging the meta-learned weight's ability to select branches without being influenced by strongly associated incorrect labels, {\ours} consistently outperforms {\oursnm} across all datasets, achieving superior performance.

We also conduct a parameter sensitivity analysis on $\alpha$ in our algorithm, which determines the influence of pseudo-labels from the multi-branch auxiliary model. Figure \ref{sensitivity} illustrates the sensitivity of {\ours} on \texttt{CIFAR-10} as $\alpha$ increases from 0.1 to 0.9. It is evident that an $\alpha$ value around 0.3 yields superior performance for {\ours}. Furthermore, we investigate the consistency and convergence of pseudo-labels generated by {\ours} on \texttt{CIFAR-10}, as shown in Figures \ref{consistency} and \ref{convergence}. It is clear that the generated pseudo-labels become consistent with the Bayes optimal classifier and converge as the number of epochs increases.

\section{Conclusion}


In this study, we aim at ID-PLL and introduce reduction-based pseudo-labels to mitigate the influence of incorrect labels during model training. Theoretically, we prove that reduction-based pseudo-labels' greater consistency with the Bayes optimal classifier. Practically, our approach achieves superior performances compared to the previous approaches, though we cost more time than other baseline due to involvement with bi-level optimization. The broader potential impact of our work includes a significant reduction in the need for accurately annotated data, thereby saving time and costs. However, this could also lead to an increase in the unemployment rate for data annotation specialists.

\bibliographystyle{unsrt}  
\bibliography{references}

\begin{thebibliography}{10}

\bibitem{cour2011learning}
Timothee Cour, Ben Sapp, and Ben Taskar.
\newblock Learning from partial labels.
\newblock {\em The Journal of Machine Learning Research}, 12:1501--1536, 2011.

\bibitem{chen2014ambiguously}
Yi-Chen Chen, Vishal~M Patel, Rama Chellappa, and P~Jonathon Phillips.
\newblock Ambiguously labeled learning using dictionaries.
\newblock {\em IEEE Transactions on Information Forensics and Security}, 9(12):2076--2088, 2014.

\bibitem{yu2016maximum}
Fei Yu and Min-Ling Zhang.
\newblock Maximum margin partial label learning.
\newblock In {\em Asian conference on machine learning}, pages 96--111. PMLR, 2016.

\bibitem{liu2012conditional}
Liping Liu and Thomas~G Dietterich.
\newblock A conditional multinomial mixture model for superset label learning.
\newblock In {\em Advances in neural information processing systems}, pages 548--556. Citeseer, 2012.

\bibitem{tang2017confidence}
Cai-Zhi Tang and Min-Ling Zhang.
\newblock Confidence-rated discriminative partial label learning.
\newblock In {\em Proceedings of the AAAI Conference on Artificial Intelligence}, volume~31, 2017.

\bibitem{luo2010learning}
Jie Luo and Francesco Orabona.
\newblock Learning from candidate labeling sets.
\newblock Technical report, MIT Press, 2010.

\bibitem{zeng2013learning}
Zinan Zeng, Shijie Xiao, Kui Jia, Tsung-Han Chan, Shenghua Gao, Dong Xu, and Yi~Ma.
\newblock Learning by associating ambiguously labeled images.
\newblock In {\em Proceedings of the IEEE Conference on Computer Vision and Pattern Recognition}, pages 708--715, 2013.

\bibitem{chen2017learning}
Ching-Hui Chen, Vishal~M Patel, and Rama Chellappa.
\newblock Learning from ambiguously labeled face images.
\newblock {\em IEEE transactions on pattern analysis and machine intelligence}, 40(7):1653--1667, 2017.

\bibitem{jin2002learning}
Rong Jin and Zoubin Ghahramani.
\newblock Learning with multiple labels.
\newblock In {\em NIPS}, volume~2, pages 897--904. Citeseer, 2002.

\bibitem{nguyen2008classification}
Nam Nguyen and Rich Caruana.
\newblock Classification with partial labels.
\newblock In {\em Proceedings of the 14th ACM SIGKDD international conference on Knowledge discovery and data mining}, pages 551--559, 2008.

\bibitem{lv2020progressive}
Jiaqi Lv, Miao Xu, Lei Feng, Gang Niu, Xin Geng, and Masashi Sugiyama.
\newblock Progressive identification of true labels for partial-label learning.
\newblock In {\em International Conference on Machine Learning}, pages 6500--6510. PMLR, 2020.

\bibitem{feng2020provably}
Lei Feng, Jiaqi Lv, Bo~Han, Miao Xu, Gang Niu, Xin Geng, Bo~An, and Masashi Sugiyama.
\newblock Provably consistent partial-label learning.
\newblock {\em arXiv preprint arXiv:2007.08929}, 2020.

\bibitem{zhang2021exploiting}
Fei Zhang, Lei Feng, Bo~Han, Tongliang Liu, Gang Niu, Tao Qin, and Masashi Sugiyama.
\newblock Exploiting class activation value for partial-label learning.
\newblock In {\em International Conference on Learning Representations}, 2021.

\bibitem{wang2022pico}
Haobo Wang, Ruixuan Xiao, Yixuan Li, Lei Feng, Gang Niu, Gang Chen, and Junbo Zhao.
\newblock Pico: Contrastive label disambiguation for partial label learning.
\newblock {\em arXiv preprint arXiv:2201.08984}, 2022.

\bibitem{wu2022revisiting}
Dong-Dong Wu, Deng-Bao Wang, and Min-Ling Zhang.
\newblock Revisiting consistency regularization for deep partial label learning.
\newblock In {\em International Conference on Machine Learning}, pages 24212--24225. PMLR, 2022.

\bibitem{hullermeier2006learning}
Eyke H{\"u}llermeier and J{\"u}rgen Beringer.
\newblock Learning from ambiguously labeled examples.
\newblock {\em Intelligent Data Analysis}, 10(5):419--439, 2006.

\bibitem{zhang2015solving}
Min-Ling Zhang and Fei Yu.
\newblock Solving the partial label learning problem: An instance-based approach.
\newblock In {\em Twenty-fourth international joint conference on artificial intelligence}, 2015.

\bibitem{lv2021robustness}
Jiaqi Lv, Biao Liu, Lei Feng, Ning Xu, Miao Xu, Bo~An, Gang Niu, Xin Geng, and Masashi Sugiyama.
\newblock On the robustness of average losses for partial-label learning.
\newblock {\em arXiv preprint arXiv:2106.06152}, 2021.

\bibitem{xu2021instance}
Ning Xu, Congyu Qiao, Xin Geng, and Min-Ling Zhang.
\newblock Instance-dependent partial label learning.
\newblock {\em Advances in Neural Information Processing Systems}, 34, 2021.

\bibitem{xia2022ambiguity}
Shiyu Xia, Jiaqi Lv, Ning Xu, and Xin Geng.
\newblock Ambiguity-induced contrastive learning for instance-dependent partial label learning.
\newblock In {\em Proceedings of the Thirty-First International Joint Conference on Artificial Intelligence, Vienna, Austria}, pages 3615--3621, 2022.

\bibitem{qiao2023decompose}
Congyu Qiao, Ning Xu, and Xin Geng.
\newblock Decompositional generation process for instance-dependent partial label learning.
\newblock In {\em The Eleventh International Conference on Learning Representations, {ICLR} 2023, Kigali, Rwanda, May 1-5, 2023}. OpenReview.net, 2023.

\bibitem{xu2023progressive}
Ning Xu, Biao Liu, Jiaqi Lv, Congyu Qiao, and Xin Geng.
\newblock Progressive purification for instance-dependent partial label learning.
\newblock In {\em International Conference on Machine Learning}, pages 38551--38565. PMLR, 2023.

\bibitem{he2023candidate}
Shuo He, Guowu Yang, and Lei Feng.
\newblock Candidate-aware selective disambiguation based on normalized entropy for instance-dependent partial-label learning.
\newblock In {\em {IEEE/CVF} International Conference on Computer Vision, Paris, France}, pages 1792--1801, 2023.

\bibitem{wu2024distilling}
Dong{-}Dong Wu, Deng{-}Bao Wang, and Min{-}Ling Zhang.
\newblock Distilling reliable knowledge for instance-dependent partial label learning.
\newblock In {\em Thirty-Eighth {AAAI} Conference on Artificial Intelligence, Vancouver, Canada}, pages 15888--15896, 2024.

\bibitem{SheHSWS021}
Jiahui She, Yibo Hu, Hailin Shi, Jun Wang, Qiu Shen, and Tao Mei.
\newblock Dive into ambiguity: Latent distribution mining and pairwise uncertainty estimation for facial expression recognition.
\newblock In {\em {IEEE} Conference on Computer Vision and Pattern Recognition, {CVPR} 2021, virtual, June 19-25, 2021}, pages 6248--6257. Computer Vision Foundation / {IEEE}, 2021.

\bibitem{zhang2016partial}
Min-Ling Zhang, Bin-Bin Zhou, and Xu-Ying Liu.
\newblock Partial label learning via feature-aware disambiguation.
\newblock In {\em Proceedings of the 22nd ACM SIGKDD international conference on knowledge discovery and data mining}, pages 1335--1344, 2016.

\bibitem{feng2018leveraging}
Lei Feng and Bo~An.
\newblock Leveraging latent label distributions for partial label learning.
\newblock In {\em IJCAI}, pages 2107--2113, 2018.

\bibitem{wang2019adaptive}
Deng-Bao Wang, Li~Li, and Min-Ling Zhang.
\newblock Adaptive graph guided disambiguation for partial label learning.
\newblock In {\em Proceedings of the 25th ACM SIGKDD International Conference on Knowledge Discovery \& Data Mining}, pages 83--91, 2019.

\bibitem{xu2019partial}
Ning Xu, Jiaqi Lv, and Xin Geng.
\newblock Partial label learning via label enhancement.
\newblock In {\em Proceedings of the AAAI Conference on Artificial Intelligence}, volume~33, pages 5557--5564, 2019.

\bibitem{yao2020deep}
Yao Yao, Jiehui Deng, Xiuhua Chen, Chen Gong, Jianxin Wu, and Jian Yang.
\newblock Deep discriminative cnn with temporal ensembling for ambiguously-labeled image classification.
\newblock In {\em Proceedings of the AAAI Conference on Artificial Intelligence}, volume~34, pages 12669--12676, 2020.

\bibitem{yao2020network}
Yao Yao, Chen Gong, Jiehui Deng, and Jian Yang.
\newblock Network cooperation with progressive disambiguation for partial label learning.
\newblock In {\em Joint European Conference on Machine Learning and Knowledge Discovery in Databases}, pages 471--488. Springer, 2020.

\bibitem{wen2021leveraged}
Hongwei Wen, Jingyi Cui, Hanyuan Hang, Jiabin Liu, Yisen Wang, and Zhouchen Lin.
\newblock Leveraged weighted loss for partial label learning.
\newblock In {\em International Conference on Machine Learning}, pages 11091--11100. PMLR, 2021.

\bibitem{he2022partial}
Shuo He, Lei Feng, Fengmao Lv, Wen Li, and Guowu Yang.
\newblock Partial label learning with semantic label representations.
\newblock In {\em Proceedings of the 28th ACM SIGKDD Conference on Knowledge Discovery and Data Mining}, pages 545--553, 2022.

\bibitem{lyu2022deep}
Gengyu Lyu, Yanan Wu, and Songhe Feng.
\newblock Deep graph matching for partial label learning.
\newblock In {\em Proceedings of the International Joint Conference on Artificial Intelligence}, pages 3306--3312, 2022.

\bibitem{tsybakov2004optimal}
Alexander~B Tsybakov.
\newblock Optimal aggregation of classifiers in statistical learning.
\newblock {\em The Annals of Statistics}, 32(1):135--166, 2004.

\bibitem{zheng2020error}
Songzhu Zheng, Pengxiang Wu, Aman Goswami, Mayank Goswami, Dimitris~N. Metaxas, and Chao Chen.
\newblock Error-bounded correction of noisy labels.
\newblock In {\em Proceedings of the 37th International Conference on Machine Learning, Virtual Event}, volume 119, pages 11447--11457, 2020.

\bibitem{shu2029meta}
Jun Shu, Qi~Xie, Lixuan Yi, Qian Zhao, Sanping Zhou, Zongben Xu, and Deyu Meng.
\newblock Meta-weight-net: Learning an explicit mapping for sample weighting.
\newblock In {\em Advances in Neural Information Processing Systems 32: Annual Conference on Neural Information Processing Systems 2019, NeurIPS 2019, December 8-14, 2019, Vancouver, BC, Canada}, pages 1917--1928, 2019.

\bibitem{paszke2019pytorch}
Adam Paszke, Sam Gross, Francisco Massa, Adam Lerer, James Bradbury, Gregory Chanan, Trevor Killeen, Zeming Lin, Natalia Gimelshein, Luca Antiga, et~al.
\newblock Pytorch: An imperative style, high-performance deep learning library.
\newblock {\em Advances in neural information processing systems}, 32, 2019.

\bibitem{krizhevsky2009learning}
Alex Krizhevsky, Geoffrey Hinton, et~al.
\newblock Learning multiple layers of features from tiny images.
\newblock 2009.

\bibitem{le2015tiny}
Ya~Le and Xuan Yang.
\newblock Tiny imagenet visual recognition challenge.
\newblock {\em CS 231N}, 7(7):3, 2015.

\bibitem{briggs2012rank}
Forrest Briggs, Xiaoli~Z Fern, and Raviv Raich.
\newblock Rank-loss support instance machines for miml instance annotation.
\newblock In {\em Proceedings of the 18th ACM SIGKDD international conference on Knowledge discovery and data mining}, pages 534--542, 2012.

\bibitem{guillaumin2010multiple}
Matthieu Guillaumin, Jakob Verbeek, and Cordelia Schmid.
\newblock Multiple instance metric learning from automatically labeled bags of faces.
\newblock In {\em European conference on computer vision}, pages 634--647. Springer, 2010.

\bibitem{liang2022efficient}
Jiajun Liang, Linze Li, Zhaodong Bing, Borui Zhao, Yao Tang, Bo~Lin, and Haoqiang Fan.
\newblock Efficient one pass self-distillation with zipf's label smoothing.
\newblock In {\em ECCV, Tel Aviv, Israel}, volume 13671, pages 104--119, 2022.

\bibitem{yin2023squeeze}
Zeyuan Yin, Eric~P. Xing, and Zhiqiang Shen.
\newblock Squeeze, recover and relabel: Dataset condensation at imagenet scale from {A} new perspective.
\newblock In {\em Annual Conference on Neural Information Processing Systems, New Orleans, LA, USA}, 2023.

\bibitem{robbins1951stochastic}
Herbert Robbins and Sutton Monro.
\newblock A stochastic approximation method.
\newblock {\em The annals of mathematical statistics}, pages 400--407, 1951.

\end{thebibliography}

\appendix
\newpage
\section{Appendix}

\subsection{Proofs of Theorem \ref{theorem1}}\label{proof:theorem1}
\textit{Proof.} For the conditional probability $p(\eta^{\star}(\bm{x}) = \arg\max_{j\in \mathcal{Y}} q^j | \bm{x}\in \mathcal{J}(\bm{x}, \bar{\mathcal{Y}}))$, we have:
\begin{equation}
    \begin{aligned}
        &p(\eta^{\star}(\bm{x}) = \arg\max_{j\in \mathcal{Y}} q^j | \bm{x}\in \mathcal{J}(\bm{x}, \bar{\mathcal{Y}})) \\
    =& \frac{p(\eta^{\star}(\bm{x}) = \arg\max_{j\in \mathcal{Y}} q^j , \bm{x}\in \mathcal{J}(\bm{x}, \bar{\mathcal{Y}}))}{p(\bm{x}\in \mathcal{J}(\bm{x}, \bar{\mathcal{Y}}))} \\
    =& \frac{p(\eta^{\star}(\bm{x}) = \arg\max_{j\in \mathcal{Y}} q^j , \eta^{\eta^{\star}(\bm{x})} - \eta^a(\bm{x}) \leq \tau, \bm{x}\in \mathcal{J}(\bm{x}, \bar{\mathcal{Y}}))}{p(\bm{x}\in \mathcal{J}(\bm{x}, \bar{\mathcal{Y}}))} \\
    &+ \frac{p(\eta^{\star}(\bm{x}) = \arg\max_{j\in \mathcal{Y}} q^j , \eta^{\eta^{\star}(\bm{x})} - \eta^a(\bm{x}) > \tau, \bm{x}\in \mathcal{J}(\bm{x}, \bar{\mathcal{Y}}))}{p(\bm{x}\in \mathcal{J}(\bm{x}, \bar{\mathcal{Y}}))} \\
    \end{aligned}
\end{equation}

Since, for $\bm{x}\in \mathcal{J}(\bm{x}, \bar{\mathcal{Y}})$, $\forall j \in \bar{\mathcal{Y}}$, $\eta^{\eta^{\star}(\bm{x})} - \eta^j(\bm{x}) \leq \tau$, we could obtain:
\begin{equation}
    \begin{aligned}
        &p(\eta^{\star}(\bm{x}) = \arg\max_{j\in \mathcal{Y}} q^j | \bm{x}\in \mathcal{J}(\bm{x}, \bar{\mathcal{Y}})) \\
        =& \frac{p(\eta^{\star}(\bm{x}) = \arg\max_{j\in \mathcal{Y}} q^j , \eta^{\eta^{\star}(\bm{x})} - \eta^a(\bm{x}) \leq \tau, \bm{x}\in \mathcal{J}(\bm{x}, \bar{\mathcal{Y}}))}{p(\bm{x}\in \mathcal{J}(\bm{x}, \bar{\mathcal{Y}}))} \\
        &+ \frac{p(\eta^{\star}(\bm{x}) = \arg\max_{j\in \mathcal{Y}} q^j , \eta^{\eta^{\star}(\bm{x})} - \eta^a(\bm{x}) > \tau, \bm{x}\in \mathcal{J}(\bm{x}, \bar{\mathcal{Y}}))}{p(\bm{x}\in \mathcal{J}(\bm{x}, \bar{\mathcal{Y}}))} \\
        =& \frac{p(\eta^{\star}(\bm{x}) = \arg\max_{j\in \mathcal{Y}} q^j , \eta^{\eta^{\star}(\bm{x})} - \eta^a(\bm{x}) \leq \tau, \bm{x}\in \mathcal{J}(\bm{x}, \bar{\mathcal{Y}}))}{p(\bm{x}\in \mathcal{J}(\bm{x}, \bar{\mathcal{Y}}))} \\
    \end{aligned}
\end{equation}

Recall that $s = \arg\max_{j\in \mathcal{Y}, j\neq \eta^{\star}(\bm{x})}\eta^{j}(\bm{x})$. since $\forall j\in \mathcal{Y}$ with $j\neq \eta^{\star}(\bm{x})$. We have 
\begin{equation}
    \begin{aligned}
        \eta^{\eta^{\star}(\bm{x})} - \eta^s(\bm{x}) \leq \eta^{\eta^{\star}(\bm{x})} - \eta^j(\bm{x}),
    \end{aligned}
\end{equation}
$\forall j \in \bar{\mathcal{Y}}$, $\eta^{\eta^{\star}(\bm{x})} - \eta^s(\bm{x}) \leq \eta^{\eta^{\star}(\bm{x})} - \eta^j(\bm{x}) \leq \tau$, and get $s = a$.

Then we could obtain:
\begin{equation}
    \begin{aligned}
        &p(\eta^{\star}(\bm{x}) = \arg\max_{j\in \mathcal{Y}} q^j | \bm{x}\in \mathcal{J}(\bm{x}, \bar{\mathcal{Y}})) \\
        =& \frac{p(\eta^{\star}(\bm{x}) = \arg\max_{j\in \mathcal{Y}} q^j , \eta^{\eta^{\star}(\bm{x})} - \eta^a(\bm{x}) \leq \tau, \bm{x}\in \mathcal{J}(\bm{x}, \bar{\mathcal{Y}}))}{p(\bm{x}\in \mathcal{J}(\bm{x}, \bar{\mathcal{Y}}))} \\
        =& \frac{p(\eta^{\star}(\bm{x}) = \arg\max_{j\in \mathcal{Y}} q^j , \eta^{\eta^{\star}(\bm{x})} - \eta^s(\bm{x}) \leq \tau, \bm{x}\in \mathcal{J}(\bm{x}, \bar{\mathcal{Y}}))}{p(\bm{x}\in \mathcal{J}(\bm{x}, \bar{\mathcal{Y}}))} \\
        =& \frac{p(\eta^{\star}(\bm{x}) = \arg\max_{j\in \mathcal{Y}} q^j , \eta^{\eta^{\star}(\bm{x})} - \eta^s(\bm{x}) \leq \tau, \eta^{\eta^{\star}(\bm{x})} - \eta^s(\bm{x}) \leq 2\epsilon, \bm{x}\in \mathcal{J}(\bm{x}, \bar{\mathcal{Y}}))}{p(\bm{x}\in \mathcal{J}(\bm{x}, \bar{\mathcal{Y}}))} \\
        &+ \frac{p(\eta^{\star}(\bm{x}) = \arg\max_{j\in \mathcal{Y}} q^j , \eta^{\eta^{\star}(\bm{x})} - \eta^s(\bm{x}) \leq \tau, \eta^{\eta^{\star}(\bm{x})} - \eta^s(\bm{x}) > 2\epsilon, \bm{x}\in \mathcal{J}(\bm{x}, \bar{\mathcal{Y}}))}{p(\bm{x}\in \mathcal{J}(\bm{x}, \bar{\mathcal{Y}}))} \\
    \end{aligned}
\end{equation}

Since $\tau \leq 2\epsilon$, we could obtain:
\begin{equation}
    \begin{aligned}
        &p(\eta^{\star}(\bm{x}) = \arg\max_{j\in \mathcal{Y}} q^j | \bm{x}\in \mathcal{J}(\bm{x}, \bar{\mathcal{Y}})) \\
        =& \frac{p(\eta^{\star}(\bm{x}) = \arg\max_{j\in \mathcal{Y}} q^j , \eta^{\eta^{\star}(\bm{x})} - \eta^s(\bm{x}) \leq \tau, \eta^{\eta^{\star}(\bm{x})} - \eta^s(\bm{x}) \leq 2\epsilon, \bm{x}\in \mathcal{J}(\bm{x}, \bar{\mathcal{Y}}))}{p(\bm{x}\in \mathcal{J}(\bm{x}, \bar{\mathcal{Y}}))} \\
        &+ \frac{p(\eta^{\star}(\bm{x}) = \arg\max_{j\in \mathcal{Y}} q^j , \eta^{\eta^{\star}(\bm{x})} - \eta^s(\bm{x}) \leq \tau, \eta^{\eta^{\star}(\bm{x})} - \eta^s(\bm{x}) > 2\epsilon, \bm{x}\in \mathcal{J}(\bm{x}, \bar{\mathcal{Y}}))}{p(\bm{x}\in \mathcal{J}(\bm{x}, \bar{\mathcal{Y}}))} \\
        =& \frac{p(\eta^{\star}(\bm{x}) = \arg\max_{j\in \mathcal{Y}} q^j , \eta^{\eta^{\star}(\bm{x})} - \eta^s(\bm{x}) \leq \tau, \eta^{\eta^{\star}(\bm{x})} - \eta^s(\bm{x}) \leq 2\epsilon, \bm{x}\in \mathcal{J}(\bm{x}, \bar{\mathcal{Y}}))}{p(\bm{x}\in \mathcal{J}(\bm{x}, \bar{\mathcal{Y}}))} \\
        \leq& \frac{p(\eta^{\star}(\bm{x}) = \arg\max_{j\in \mathcal{Y}} q^j , \eta^{\eta^{\star}(\bm{x})} - \eta^s(\bm{x}) \leq 2\epsilon, \bm{x}\in \mathcal{J}(\bm{x}, \bar{\mathcal{Y}}))}{p(\bm{x}\in \mathcal{J}(\bm{x}, \bar{\mathcal{Y}}))} \\
        \leq& \frac{p(\eta^{\eta^{\star}(\bm{x})} - \eta^s(\bm{x}) \leq 2\epsilon, \bm{x}\in \mathcal{J}(\bm{x}, \bar{\mathcal{Y}}))}{p(\bm{x}\in \mathcal{J}(\bm{x}, \bar{\mathcal{Y}}))} \\
    \end{aligned}
\end{equation}

Recall that $\epsilon' \in (0, \min \{1, \min_{\bm{x}\in \mathcal{J}(\bm{x}, \bar{\mathcal{Y}})} \frac{(\eta^{\eta^{\star}(\bm{x})}(\bm{x}) - \eta^b(\bm{x}))(\eta^{\eta^{\star}(\bm{x})}(\bm{x}) -\eta^a(\bm{x}))}{4\epsilon(1 - \sum_{j \in \bar{\mathcal{Y}}}\eta^j(\bm{x}))}\})$ with $a = \arg\max_{j\in \bar{\mathcal{Y}}}\eta^{j}(\bm{x})$ and $b = \arg\max_{j\notin \{y\}\cup\bar{\mathcal{Y}}}\eta^{j}(\bm{x})$. We have 
\begin{equation}
    \begin{aligned}
        \frac{(\eta^{\eta^{\star}(\bm{x})}(\bm{x}) - \eta^b(\bm{x}))(\eta^{\eta^{\star}(\bm{x})}(\bm{x}) -\eta^a(\bm{x}))}{4\epsilon(1 - \sum_{j \in \bar{\mathcal{Y}}}\eta^j(\bm{x}))} \geq 2\epsilon' \\
        \frac{(\eta^{\eta^{\star}(\bm{x})}(\bm{x}) - \eta^b(\bm{x}))(\eta^{\eta^{\star}(\bm{x})}(\bm{x}) -\eta^s(\bm{x}))}{4\epsilon(1 - \sum_{j \in \bar{\mathcal{Y}}}\eta^j(\bm{x}))} \geq 2\epsilon' \\
        \frac{(\eta^{\eta^{\star}(\bm{x})}(\bm{x}) - \eta^b(\bm{x}))2\epsilon}{4\epsilon(1 - \sum_{j \in \bar{\mathcal{Y}}}\eta^j(\bm{x}))} \geq 2\epsilon' \\
        \frac{\eta^{\eta^{\star}(\bm{x})}(\bm{x}) - \eta^b(\bm{x})}{1 - \sum_{j \in \bar{\mathcal{Y}}}\eta^j(\bm{x})} \geq 2\epsilon' \\
    \end{aligned}
\end{equation}

Here, according to Eq. (\ref{eta'}), we have $\eta'^{\eta'^{\star}(\bm{x})} - \eta'^s(\bm{x}) > 2\epsilon'$, and obtain:

\begin{equation}
    \begin{aligned}
        &p(\eta^{\star}(\bm{x}) = \arg\max_{j\in \mathcal{Y}} q^j | \bm{x}\in \mathcal{J}(\bm{x}, \bar{\mathcal{Y}})) \\
        \leq& \frac{p(\eta^{\eta^{\star}(\bm{x})} - \eta^s(\bm{x}) \leq 2\epsilon, \bm{x}\in \mathcal{J}(\bm{x}, \bar{\mathcal{Y}}))}{p(\bm{x}\in \mathcal{J}(\bm{x}, \bar{\mathcal{Y}}))} \\
        \leq & \frac{p(\eta'^{\eta'^{\star}(\bm{x})} - \eta'^s(\bm{x}) > 2\epsilon', \bm{x}\in \mathcal{J}(\bm{x}, \bar{\mathcal{Y}}))}{p(\bm{x}\in \mathcal{J}(\bm{x}, \bar{\mathcal{Y}}))} \\
    \end{aligned}
\end{equation}

Recall that $\forall j \in \mathcal{Y}\setminus \bar{\mathcal{Y}}$, $| \varphi^j(\bm{x}) - \eta'^j(\bm{x}) \leq \epsilon'$. We have
\begin{equation}
    \begin{aligned}
        \eta'^{\eta'^{\star}(\bm{x})} - \eta'^s(\bm{x}) &> 2\epsilon' \\
        \eta'^{\eta'^{\star}(\bm{x})} - \epsilon' &>  \eta'^s(\bm{x}) + \epsilon' \\
        \eta'^{\eta'^{\star}(\bm{x})} - \epsilon' &>  \eta'^j(\bm{x}) + \epsilon', \forall j\in \mathcal{Y}, j\neq \eta'^{\star}(\bm{x}) \\
        \varphi^{\eta'^{\star}}(\bm{x}) &> \varphi^j(\bm{x}), \forall j\in \mathcal{Y}, j\neq \eta'^{\star}(\bm{x}) \\
    \end{aligned}
\end{equation}

Take $\arg\max_{j\in\mathcal{Y}} q^{j} = \arg\max_{j\in\mathcal{Y}} \varphi^{j}(\bm{x})$, we could obtain $\eta^{\star}(\bm{x}) = \arg\max_{j\in \mathcal{Y}} q'^j $.

Finally, we have
\begin{equation}
    \begin{aligned}
    &p(\eta^{\star}(\bm{x}) = \arg\max_{j\in \mathcal{Y}} q^j | \bm{x}\in \mathcal{J}(\bm{x}, \bar{\mathcal{Y}})) \\
    =& \frac{p(\eta^{\star}(\bm{x}) = \arg\max_{j\in \mathcal{Y}} q^j , \bm{x}\in \mathcal{J}(\bm{x}, \bar{\mathcal{Y}}))}{p(\bm{x}\in \mathcal{J}(\bm{x}, \bar{\mathcal{Y}}))} \\
    =& \frac{p(\eta^{\star}(\bm{x}) = \arg\max_{j\in \mathcal{Y}} q^j , \eta^{\eta^{\star}(\bm{x})} - \eta^a(\bm{x}) \leq \tau, \bm{x}\in \mathcal{J}(\bm{x}, \bar{\mathcal{Y}}))}{p(\bm{x}\in \mathcal{J}(\bm{x}, \bar{\mathcal{Y}}))} \\
    &+ \frac{p(\eta^{\star}(\bm{x}) = \arg\max_{j\in \mathcal{Y}} q^j , \eta^{\eta^{\star}(\bm{x})} - \eta^a(\bm{x}) > \tau, \bm{x}\in \mathcal{J}(\bm{x}, \bar{\mathcal{Y}}))}{p(\bm{x}\in \mathcal{J}(\bm{x}, \bar{\mathcal{Y}}))} \\
    =& \frac{p(\eta^{\star}(\bm{x}) = \arg\max_{j\in \mathcal{Y}} q^j , \eta^{\eta^{\star}(\bm{x})} - \eta^a(\bm{x}) \leq \tau, \bm{x}\in \mathcal{J}(\bm{x}, \bar{\mathcal{Y}}))}{p(\bm{x}\in \mathcal{J}(\bm{x}, \bar{\mathcal{Y}}))} \\
    =& \frac{p(\eta^{\star}(\bm{x}) = \arg\max_{j\in \mathcal{Y}} q^j , \eta^{\eta^{\star}(\bm{x})} - \eta^s(\bm{x}) \leq \tau, \bm{x}\in \mathcal{J}(\bm{x}, \bar{\mathcal{Y}}))}{p(\bm{x}\in \mathcal{J}(\bm{x}, \bar{\mathcal{Y}}))} \\
    =& \frac{p(\eta^{\star}(\bm{x}) = \arg\max_{j\in \mathcal{Y}} q^j , \eta^{\eta^{\star}(\bm{x})} - \eta^s(\bm{x}) \leq \tau, \eta^{\eta^{\star}(\bm{x})} - \eta^s(\bm{x}) \leq 2\epsilon, \bm{x}\in \mathcal{J}(\bm{x}, \bar{\mathcal{Y}}))}{p(\bm{x}\in \mathcal{J}(\bm{x}, \bar{\mathcal{Y}}))} \\
    &+ \frac{p(\eta^{\star}(\bm{x}) = \arg\max_{j\in \mathcal{Y}} q^j , \eta^{\eta^{\star}(\bm{x})} - \eta^s(\bm{x}) \leq \tau, \eta^{\eta^{\star}(\bm{x})} - \eta^s(\bm{x}) > 2\epsilon, \bm{x}\in \mathcal{J}(\bm{x}, \bar{\mathcal{Y}}))}{p(\bm{x}\in \mathcal{J}(\bm{x}, \bar{\mathcal{Y}}))} \\
    =& \frac{p(\eta^{\star}(\bm{x}) = \arg\max_{j\in \mathcal{Y}} q^j , \eta^{\eta^{\star}(\bm{x})} - \eta^s(\bm{x}) \leq \tau, \eta^{\eta^{\star}(\bm{x})} - \eta^s(\bm{x}) \leq 2\epsilon, \bm{x}\in \mathcal{J}(\bm{x}, \bar{\mathcal{Y}}))}{p(\bm{x}\in \mathcal{J}(\bm{x}, \bar{\mathcal{Y}}))} \\
    \leq& \frac{p(\eta^{\star}(\bm{x}) = \arg\max_{j\in \mathcal{Y}} q^j , \eta^{\eta^{\star}(\bm{x})} - \eta^s(\bm{x}) \leq 2\epsilon, \bm{x}\in \mathcal{J}(\bm{x}, \bar{\mathcal{Y}}))}{p(\bm{x}\in \mathcal{J}(\bm{x}, \bar{\mathcal{Y}}))} \\
    \leq& \frac{p(\eta^{\eta^{\star}(\bm{x})} - \eta^s(\bm{x}) \leq 2\epsilon, \bm{x}\in \mathcal{J}(\bm{x}, \bar{\mathcal{Y}}))}{p(\bm{x}\in \mathcal{J}(\bm{x}, \bar{\mathcal{Y}}))} \\
    \leq & \frac{p(\eta'^{\eta'^{\star}(\bm{x})} - \eta'^s(\bm{x}) > 2\epsilon', \bm{x}\in \mathcal{J}(\bm{x}, \bar{\mathcal{Y}}))}{p(\bm{x}\in \mathcal{J}(\bm{x}, \bar{\mathcal{Y}}))} \\
    \leq & \frac{p(\eta^{\star}(\bm{x}) = \arg\max_{j\in \mathcal{Y}} q'^j, \bm{x}\in \mathcal{J}(\bm{x}, \bar{\mathcal{Y}}))}{p(\bm{x}\in \mathcal{J}(\bm{x}, \bar{\mathcal{Y}}))} \\
    =& p(\eta^{\star}(\bm{x}) = \arg\max_{j\in \mathcal{Y}} q'^j | \bm{x}\in \mathcal{J}(\bm{x}, \bar{\mathcal{Y}}))
    \end{aligned}
\end{equation}

The proof has been completed.

\subsection{Proofs of Theorem \ref{theorem2}}\label{proof:theorem2}
\begin{equation}
    \begin{aligned}
        & p(\eta^{\star}(\bm{x}) = \arg\max_{j\in \mathcal{Y}} q'^j | \bm{x}\in \mathcal{J}(\bm{x}, \bar{\mathcal{Y}})) \\
    \geq& p(\eta'^{\eta'^{\star}(\bm{x})} - \eta'^s(\bm{x}) > 2\epsilon' | \bm{x}\in \mathcal{J}(\bm{x}, \bar{\mathcal{Y}})) \\
    \geq& p(\eta^{\eta^{\star}(\bm{x})} - \eta^s(\bm{x}) > \frac{4\epsilon\epsilon'(1 - \eta^s(\bm{x}))}{\eta^{\eta^{\star}(\bm{x})} - \eta^b(\bm{x})} | \bm{x}\in \mathcal{J}(\bm{x}, \bar{\mathcal{Y}})) \\
    =& 1 - p(\eta^{\eta^{\star}(\bm{x})} - \eta^s(\bm{x}) \leq \frac{4\epsilon\epsilon'(1 - \eta^s(\bm{x}))}{\eta^{\eta^{\star}(\bm{x})} - \eta^b(\bm{x})} | \bm{x}\in \mathcal{J}(\bm{x}, \bar{\mathcal{Y}})) \\
    \end{aligned}     
\end{equation}

Since  for $\bm{x} \in J(\bm{x}, \bar{\mathcal{Y}})$, its posterior $\eta(\bm{x})$ fulfills Assumption \ref{Tsybakov} for constants $C, \lambda > 0$ and $t_0 \in (0, 1]$, we have
\begin{equation}
    \begin{aligned}
        & p(\eta^{\star}(\bm{x}) = \arg\max_{j\in \mathcal{Y}} q'^j | \bm{x}\in \mathcal{J}(\bm{x}, \bar{\mathcal{Y}})) \\
        \geq& 1 - C(\frac{4\epsilon\epsilon'(1 - \eta^s(\bm{x}))}{\eta^{\eta^{\star}(\bm{x})} - \eta^b(\bm{x})})^{\lambda} \\
        =& 1 - C[O(\epsilon\epsilon')]^{\lambda}
    \end{aligned}
\end{equation}

\end{document}